\documentclass[10pt,twocolumn,letterpaper]{article}

\usepackage[pagenumbers]{cvpr} 

\usepackage{graphicx}
\usepackage{amsmath}
\usepackage{amssymb}
\usepackage{booktabs}

\usepackage[pagebackref,breaklinks,colorlinks]{hyperref}

\usepackage{gensymb}
\usepackage{amsfonts,amssymb}
\usepackage{pifont}
\newcommand{\cmark}{\textcolor{cyan}{\ding{51}}}
\newcommand{\xmark}{\textcolor{red}{\ding{55}}}
\usepackage{makecell, booktabs}
\usepackage{graphicx}
\usepackage{amsmath}

\usepackage[accsupp]{axessibility}

\usepackage[capitalize]{cleveref}
\crefname{section}{Sec.}{Secs.}
\Crefname{section}{Section}{Sections}
\Crefname{table}{Table}{Tables}
\crefname{table}{Tab.}{Tabs.}

\def\cvprPaperID{11986} 
\def\confName{CVPR}
\def\confYear{2023}
\newcommand{\methodname}{\textsc{VLN-SIG}}

\begin{document}

\title{Improving Vision-and-Language Navigation by Generating \\ Future-View Image Semantics}

\author{Jialu Li \quad \quad Mohit Bansal
 \\ 
   UNC Chapel Hill\\ 
   \texttt{\{jialuli, mbansal\}@cs.unc.edu} \\
   {\tt \normalsize \href{https://jialuli-luka.github.io/VLN-SIG}{https://jialuli-luka.github.io/VLN-SIG}}
}
\maketitle

\begin{abstract}
   Vision-and-Language Navigation (VLN) is the task that requires an agent to navigate through the environment based on natural language instructions. 
   At each step, the agent takes the next action by selecting from a set of navigable locations. In this paper, we aim to take one step further and explore whether the agent can benefit from generating the potential future view during navigation. Intuitively, humans will have an expectation of how the future environment will look like, based on the natural language instructions and surrounding views, which will aid correct navigation. Hence, to equip the agent with this ability to generate the semantics of future navigation views, we first propose three proxy tasks during the agent's in-domain pre-training: Masked Panorama Modeling (MPM), Masked Trajectory Modeling (MTM), and Action Prediction with Image Generation (APIG). These three objectives teach the model to predict missing views in a panorama (MPM), predict missing steps in the full trajectory (MTM), and generate the next view based on the full instruction and navigation history (APIG), respectively. 
We then fine-tune the agent on the VLN task with an auxiliary loss that minimizes the difference between the view semantics generated by the agent and the ground truth view semantics of the next step. Empirically, our \methodname{} achieves the new state-of-the-art on both Room-to-Room dataset and CVDN dataset. We further show that our agent learns to fill in missing patches in future views qualitatively, which brings more interpretability over agents' predicted actions. Lastly, we demonstrate that learning to predict future view semantics also enables the agent to have better performance on longer paths.
\end{abstract}

\section{Introduction}

\begin{figure}[t]
\begin{center}
\includegraphics[width=0.9\linewidth]{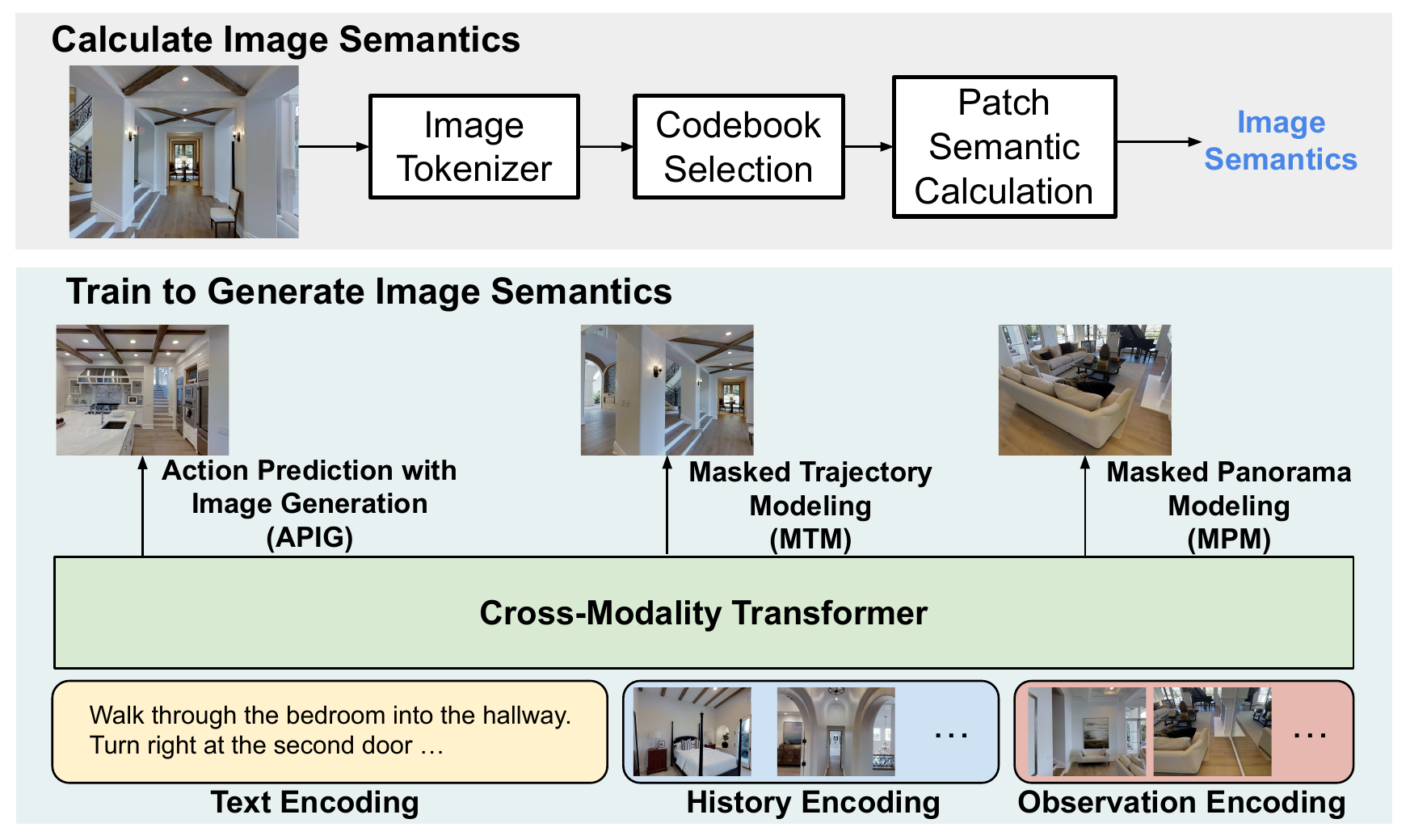}
\end{center}
\vspace{-10pt}
  \caption{Overview of our proposed method \methodname{}. We obtain the semantics of an image with a pre-trained image tokenizer, and use codebook selection (Sec.~\ref{sec:class_selection}) and patch semantic calculation (Sec.~\ref{sec:mean_sample_weight}) to adapt it to efficient in-domain VLN learning. We pre-train the agent on three proxy tasks (Sec.~\ref{sec:pt_task}) and fine-tune the agent using Action Prediction with Image Generation (APIG) as the additional auxiliary task (Sec.~\ref{sec:ft}). 
}
\vspace{-20pt}
\label{figure1}
\end{figure}

In Vision-and-Language Navigation, the agent needs to navigate through the environment based on natural language instructions. Many datasets have been proposed to solve this challenging task \cite{anderson2018vision, ku2020room, qi2020reverie, thomason2020vision, chen2019touchdown}. 
Based on these datasets, previous works aim to strengthen the navigation model or agent from several aspects: understanding long and detailed instructions in different languages, inferring and locating target objects based on common knowledge, navigating in diverse and potentially unseen environments, and learning better alignment between the environment and the instructions. However, most previous work simplifies the navigation process as an action selection process, where at each time step, the agent picks the action from a set of pre-defined candidates. 
This simplification does not take advantage of human's ability to expect what scene will be more likely to happen next during navigation. For example, given the instruction ``Walk through the bedroom into the hallway. Turn right and wait in the kitchen doorway.", before walking out of the bedroom, humans will expect the hallway to be a long passage with doors, and the kitchen probably contains objects like a kitchen island and sink. Humans have these expectations based on common sense knowledge and use them to select candidates during navigation. Thus, in this paper, we explore whether AI agents could also benefit from the ability to generate future scenes for action selection during navigation. 

\cite{koh2021pathdreamer} first explores the useful idea of generating future scenes with high quality, based on history observations on an indoor Vision-and-Language Navigation dataset. Furthermore, they adapt their synthetic future scenes for VLN by replacing the original observations with the generated observations. However, 
their replacement method did not enhance the agents' performance on the VLN task. The potential of using semantic information from generated future observations is still underexplored. Thus, in this paper, we aim to equip the agent with both the ability to predict future scenes and also benefit from learning semantics in future generated observations.

We propose \methodname{}: Vision-and-Language Navigation with Image Semantics Generation. As shown in Figure~\ref{figure1}, to first calculate the overall image semantics, we tokenize the view images into visual tokens with a pre-trained discrete variational autoencoder (dVAE). While the pre-trained dVAE has a large vocabulary of 8192, we propose and compare static and dynamic codebook selection processes to optimize a subset of the codebook at one time during training. This helps the agent to learn the more important tokens and focus on optimizing more difficult tokens. Then, we propose and compare three ways based on mean value, block-wise weighted value, and sampling to represent the semantics of all patches for efficient image semantics learning.
In the pre-training stage, we propose three tasks: Masked Panorama Modeling (MPM), Masked Trajectory Modeling (MTM) and Action Prediction with Image Generation (APIG). In Masked Panorama Modeling, the agent learns to predict the semantics of multiple missing views in a full panorama. In Masked Trajectory Modeling, the agent learns to predict the semantics of multiple steps in the full trajectory. MPM and MTM together give the agent the ability to understand the semantics in each visual token and learn to recognize the semantics contained in each view. In Action Prediction with Image Generation, the agent mimics the navigation process to predict the next step by generating the visual tokens contained in the next navigation view. This task enables the agent to imagine the next step view semantics before making actions. 
We then fine-tune the agent on the step-by-step Vision-and-Language Navigation task. We further enhance agents' ability to predict future views by optimizing an auxiliary task during navigation, which minimizes the difference between predicted observations and the target observation. Though this task does not help the agent make navigation decision directly, the future visual semantics injected by this task helps the agent understand the environment better.

We conduct experiments on Room-to-Room (R2R) datasets \cite{anderson2018vision} and Cooperative Vision-and-Dialog Navigation (CVDN) dataset~\cite{thomason2020vision}. Empirical results show that our proposed \methodname{} outperforms the strong SotA agents~\cite{chen2021history, chen2022think} by a relative gain of 4.5\% in goal progress (meters) on CVDN test leaderboard, and 3\% absolute gain in success rate on Room-to-Room test leaderboard. We further demonstrate that our proposed codebook selection methods and patch semantic calculation methods are crucial for learning to generate image semantics with ablation studies. Besides, we show that our agent achieves better performance for longer paths. Lastly, we show that our agent learns to fill in missing patches in the future views, which brings more interpretability over agents' predictions.

\section{Related Work}
\noindent\textbf{Vision-and-Language Navigation.} Vision-and-Language Navigation requires an agent to navigate based on natural language instructions. The navigation environment can be indoor~\cite{anderson2018vision, ku2020room, qi2020reverie, thomason2020vision}, outdoor~\cite{chen2019touchdown, mehta2020retouchdown} or synthetic~\cite{macmahon2006walk, misra2018mapping, shridhar2020alfred}. Previous work improves agents' navigation performance from several aspects. \cite{lin2022adapt, zhang2022explicit, li2021improving, raychaudhuri2021language, wang2019reinforced, zhang2022explicit, landi2021multimodal, hong2020language} aim to learn better alignment between language, environment and actions, while \cite{tan2019learning, wang2020environment, li2022envedit} improve agents' generalization to unseen environments. Besides, graph information is incorporated as additional signals for pre-exploration or back-tracking to aid navigation ~\cite{chen2022think, zhu2021soon, deng2020evolving, georgakis2022cross, chen2021topological, chen2022weakly}. Furthermore, pre-training techniques have been widely applied in VLN tasks, where \cite{majumdar2020improving, hao2020towards, hong2020recurrent, chen2021history, guhur2021airbert, qiao2022hop, chen2022learning} adapt pre-trained cross-modal representations or perform in-domain VLN pretraining to learn a better uni-modal and cross-modal representations. Different from previous agents which make action solely by picking from a set of candidates, we aim to equip the agent with the ability to generate semantics in future views to aid decision making. \cite{koh2021pathdreamer} proposes a separate image generation model to generate future scenes based on history observation in VLN. However, they use the generated future scenes as a substitution for the original environments, and their agent does not benefit from learning to navigate given generated scenes. Differently, we propose three new pre-training tasks that guide the agent to learn and generate the visual semantics in the observations, and further enhance the agent's ability to imagine the next step view semantics by optimizing an auxiliary task during fine-tuning.

\noindent\textbf{Vision-and-Language Pre-training.} Pre-training has been a popular technique to learn better uni-modal or multi-modal representations from large datasets. Many tasks have been proposed. For example, Masked Language Modeling (MLM)~\cite{devlin2018bert, liu2019roberta} is used for language-only pre-training, and Masked Region Modeling (MRM), and Image Text Matching (ITM)~\cite{tan2019lxmert, lu2019vilbert} have been proposed for Vision-and-Language pre-training. Recently, \cite{bao2021beit} proposes Masked Image Modeling (MIM) for vision-only pre-training, where they mask several patches in an image and predict their corresponding visual tokens. MIM has been extended successfully to Vision-and-Language pre-training~\cite{bao2022vl} and Video-and-Language pre-training~\cite{ge2022miles}. Inspired by them, we propose three VLN specific tasks based on MIM. To effectively and efficiently adapt MIM to VLN pre-training, we change the task objective to mask full images instead of patches in the image, and propose two ways for visual token codebook selection and three methods to represent overall image semantics.

\section{Method}

\begin{figure*}[t]
\begin{center}
\includegraphics[width=0.95\linewidth]{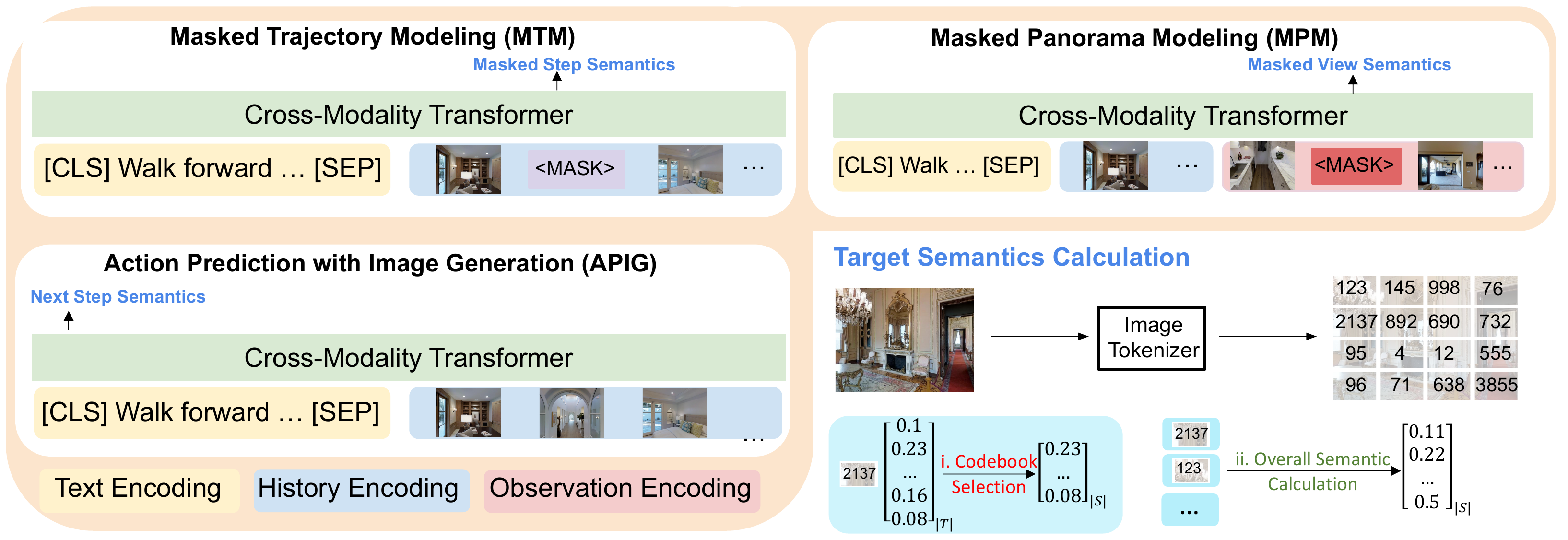}
\end{center}
\vspace{-15pt}
  \caption{Illustration of the target semantics calculation process and our proposed three proxy tasks.
}
\label{figure2}
\vspace{-10pt}
\end{figure*}

\subsection{Problem Setup}

Vision-and-Language Navigation (VLN) requires an agent to navigate through the environment based on natural language instructions. Formally, the agent navigates from start location $B$ towards target location $C$ based on language instructions $I$. At each navigation step, the agent has access to the panorama view $P$ of the current location, which is discretized into 36 view images, and the agent will pick the next step from a set of $K$ navigable locations $\{C_i\}_{i=1}^{K}$. The agent should predict a ``STOP" action when it reaches the target location $C$.

\subsection{Pre-training Overview} \label{sec:pretrain}
In the pre-training stage, the agent is pre-trained on six proxy tasks as in ~\cite{chen2021history} and the three proxy tasks we propose in this paper. Specifically, the six tasks we adopt are Masked Language Modeling (MLM), Masked Region Modeling (MRM), Instruction Trajectory Matching (ITM), Single-step Action Prediction/Regression (SAP/SAR), and Spatial Relationship Prediction (SPREL). 
Details of these tasks can be found in Appendix. 

Besides these, we further propose three tasks to help the agent learn to generate image semantics based on multi-modal information to aid decision making during navigation (Figure~\ref{figure2}).
Formally, given an image $v_i\in \mathbb{R}^{H\times W\times C}$, 
we tokenize the image into $N$ visual tokens $\{m_{ij}\}_{j=1}^{N}$ with an image tokenizer, where each visual token $m_{ij}$ belongs to the vocabulary $T=\{1,...,|T|\}$, and $N=HW/P^2$, $P$ is the patch size. Following \cite{bao2021beit}, we adopt the pre-trained discrete variational autoencoder (dVAE) from \cite{ramesh2021zero} to compute the probability $p_{ij}\in \mathbb{R}^{|T|}$ ($\sum_{k=1}^{|T|}p_{ijk} = 1$) for each visual token:
\begin{equation}
    p_{ij1}, ..., p_{ij|T|} = \mathrm{dVAE\_Enc}(v_i)
\end{equation}

\paragraph{Improving Efficiency by Generating Overall Image Semantics.} Each visual token contains specific semantic information in the image. In Masked Image Modeling used in previous vision transformer pre-training~\cite{bao2021beit, bao2022vl}, the model learns to predict the visual token for each masked patch in an image to learn the semantics for each patch. Different from them, we directly mask the full image and predict the overall semantics of the image. Our approach is \textit{better in both memory cost and speed} for VLN tasks. 
First, 
by masking the full image, we are able to directly use the pre-extracted features and eliminate the need to tune the visual backbone that encodes images with masked patches. Our approach is more than two times faster than \cite{chen2021history}, which tunes the visual backbone end-to-end to enhance the agents' performance.
Second, directly masking the full image and generating the overall semantics saves memory. It does not need to load the visual backbone in memory, and it takes the image representation as input, shrinking the input sequence length from $N*A$ (since previous MIM work \cite{bao2021beit} takes the patch representation as input) to $A$, where $A$ is the sequence length of navigation history, and $N$ is the number of patches in one image. To use our approach, it's important to have a good representation of the overall semantics of the image as the target. Thus, we explore and compare three ways to represent the probability of generating the overall semantics of the image in Sec.~\ref{sec:mean_sample_weight}. 

Furthermore, since the discrete variational autoencoder is learned from a large amount of text-image pairs collected from the internet, not every learned visual token in the vocabulary is of equal importance for the views in the Vision-and-Language Navigation environment. Thus, we propose two ways to optimize over a subset of visual tokens $S \subseteq T$ in Sec.~\ref{sec:class_selection}. 

Finally, we describe our three newly proposed pre-training proxy tasks that help the agent learn to generate the semantics of an image Sec.~\ref{sec:pt_task}.

\subsection{Overall Image Semantics Generation} \label{sec:mean_sample_weight}

In Vision-and-Language Navigation pre-training, the vision-modality input contains navigation history, which is represented as a sequence of $A$ views. If we predict the generation of all masked patches in one image, it results in an input sequence length of $A*N$, which makes it computationally infeasible for VLN pre-training. 
Thus, we propose and compare three ways to efficiently predict overall image semantics generation probabilities.

\paragraph{Mean Patch Probability.} 
The easiest way to represent the overall generation probability is to take the mean of all patches' probabilities. Specifically, given an image $v_i$, we represent the generation probability $p^{o_m}_{ik}$ as:

\begin{equation}
    p^{o_m}_{ik} = \frac{1}{N}\sum_{j=1}^{N}{p_{ijk}}
\end{equation}

where $o$ indicates \textbf{o}verall generation probability and $m$ indicates \textbf{m}ean patch probability.

\paragraph{Sample Patch Probability.}
Using the mean patch probability has one main drawback: when we optimize the KL-divergence loss between the predicted probability $\hat{p}^{o_m}_{ik}$ and the ground truth probability $p^{o_m}_{ik}$, it does not guarantee that the agent learns the semantics of every patch ($p_{ijk}$) correctly. Formally, after optimization, our goal is that for every patch $j$ in image $v_i$, $p_{ijk} - \hat{p}_{ijk} < \epsilon $, where $\epsilon$ is as small as possible. Directly optimizing the mean patch probability only minimizes the difference between $p^{o_m}_{ik}$ ($\frac{1}{N}\sum_{j=1}^{N}{p_{ijk}}$) and $\hat{p}^{o_m}_{ik}$ ($\frac{1}{N}\sum_{j=1}^{N}{\hat{p}_{ijk}}$), but for each patch $j$, $p_{ijk}$ and $\hat{p}_{ijk}$ can still be different. Besides, as the model only learns to predict the mean patch probability, it cannot actually predict the token for each patch and generate the real images. To mitigate this problem, we propose sample patch probability. We sample one patch from $N$ patches in an image, and use the semantics in the sampled patch to represent the image:

\begin{equation}
     p^{o_s}_{ik} = p_{ijk}
\end{equation}
where $j$ is the sampled patch.
We predict $p^{o_s}_{ik}$ based on encoded image inputs and sampled patch $j$'s position. We encode the position of the patch with a learned position embedding, and add it to the encoded inputs.

\paragraph{Weighted Patch Probability.}
Sample patch probability aims to minimize the differences between $p_{ijk}$ and $\hat{p}_{ijk}$ for every patch $j$ in the image. However, minimizing the difference between $p_{ij_1k}$ and $\hat{p}_{ij_1k}$ for patch $j_1$ might push $p_{ij_2k}$ and $\hat{p}_{ij_2k}$ away for patch $j_2$, which makes the convergence harder. Thus, we further propose weighted patch probability. 
Formally, we represent image semantics generation probability $p^{o_w}_{ik}$ as:
\begin{equation}
    p^{o_w}_{ik} = \sum_{j=1}^{N}{w_jp_{ijk}} 
\end{equation}
where $w_j$ is a randomly sampled weight for patch $j$, and $\sum_{j=1}^N{w_j}=1$. Since we randomly sample $\{w_j\}_{j=1}^N$ during training, after optimization, $\forall \{w_j\}_{j=1}^N$, $p^{o_w}_{ik} - \hat{p}^{o_w}_{ik} < \epsilon$. This constraint guarantees that for every patch $j$ in the image $v_i$, $p_{ijk} - \hat{p}_{ijk} < \epsilon$ after optimization. 
Detailed proof can be found in Appendix. Furthermore, we propose block-wise weighted patch probability, where at each optimization step, we randomly sample a block which contains $B$ patches, and the weights for the patches in the block are higher than other patches. The block-wise weights help the model learn to predict the richer semantics in the block compared with single token. 
We predict the weighted patch probability based on encoded image inputs and sampled weights, where we encode the sampled weights $\{w_j\}_{j=1}^N$ with fully-connected layers.

\subsection{Visual Token Codebook Selection} \label{sec:class_selection}

Since the discrete variational autoencoder (dVAE)~\cite{ramesh2021zero} is trained on image-text pairs from the web, not all the 8192 classes are of equal importance for the environment in VLN. Besides, it's hard to adapt 8192 classes to the VLN domain at the same time. Thus, we propose two ways to learn a subset of the vocabulary at each iteration during optimization. 

\paragraph{Static Codebook Selection.}

We first propose a simple static codebook selection method, where we select a fixed set of $|S|$ visual tokens from the vocabulary $T$. Specifically, we calculate the visual token frequency in the training environment, and pick the $|S|$ visual tokens that have the highest frequency. During training, we only predict the generation probability of these static $|S|$ visual tokens. This static codebook selection filters out tokens that are rarely appeared in the dataset. Thus the agent could focus more on the frequent tokens.

\paragraph{Dynamic Codebook Selection.}

We propose a novel dynamic codebook selection method, where we dynamically select a set of $S$ visual tokens from the vocabulary $T$. Specifically, the $S$ tokens are initialized to be the tokens that have the highest frequency in the environments. During training, we update the frequency score $s_f$ based on visual tokens that appear in the current training batch. Besides, we add a difficulty score $s_d$ based on the difference of the agents' predicted probability $\hat{p}^o_{ik}$ and the ground truth probability $p^o_{ik}$ ($p^o_{ik}\in \{p^{o_m}_{ik}, p^{o_s}_{ik}, p^{o_w}_{ik}\}$). We pick the $S$ tokens that have the highest score $s_{t}$ at training step $t$:
\begin{align}
    s_d &= |\hat{p}^o_{ik} - p^o_{ik}| \\ 
    s_{t} &= \lambda s_{t-1} + (1-\lambda)(s_f + \gamma s_d)
\end{align}
The score $s_{t}$ takes both token frequency and learning difficulty into consideration. The agent learns to focus on more frequent visual tokens and also puts more effort into learning the more difficult tokens.

\subsection{Pre-training Tasks} \label{sec:pt_task}
We propose three proxy tasks for the agent to learn to generate image semantics calculated with the codebook selection (Sec.~\ref{sec:class_selection}) and patch semantic calculation (Sec.~\ref{sec:mean_sample_weight}). We directly adopt the history-aware transformer model HAMT~\cite{chen2021history} as our base agent, and pre-train it from scratch with the 6 tasks proposed by \cite{chen2021history} and our proposed 3 tasks.

\begin{table*}
    \centering
     \resizebox{1.1\columnwidth}{!}{
    \begin{tabular}{|c|cc|cc|c|c|}
    \hline 
     \textbf{Model}  & \multicolumn{4}{c|}{\textbf{R2R}}  &
       \multicolumn{2}{c|}{\textbf{CVDN}}\\  \hline 
     &  \multicolumn{2}{c|}{\textbf{Val Unseen}} & \multicolumn{2}{c|}{\textbf{Test}} &  \multicolumn{1}{c|}{\textbf{Val Unseen}} & \multicolumn{1}{c|}{\textbf{Test}}\\ \hline
       & \textbf{SR$\uparrow$} & \textbf{SPL$\uparrow$} & \textbf{SR$\uparrow$} & \textbf{SPL$\uparrow$} & \textbf{GP$\uparrow$} & \textbf{GP$\uparrow$} \\ \hline
PREVALENT~\cite{hao2020towards} & 58 & 53 & 54 & 51 & 3.15 & 2.44 \\
$\circlearrowright$BERT~\cite{hong2020recurrent} & 63 & 57 & 63 & 57 & - & - \\ 
HOP~\cite{qiao2022hop} & 64 & 57 & 64 & 59 & 4.41&  3.31 \\ 
HAMT~\cite{chen2021history}  & 66 & 61 & 65 & 60 & 5.13 & 5.58 \\
Ours &  \textbf{68} & \textbf{62} & \textbf{65} & \textbf{60} &\textbf{5.52} & \textbf{5.83} \\ \hline 
SSM~\cite{wang2021structured}$^{\spadesuit}$ & 62 & 45 & 61 & 46 & - & - \\
DUET~\cite{chen2022think}$^{\spadesuit}$ & 72 & 60 & 69 & 59 & - & - \\ 
Ours$^{\spadesuit}$ & \textbf{72} & \textbf{62} & \textbf{72} & \textbf{60} & - & -\\
    \hline
    \end{tabular}
    }
    \vspace{-5pt}
    \caption{Comparison with state-of-the-art agents on Room-to-Room (R2R) and Cooperative Vision-and-Dialog Navigation (CVDN) validation unseen set and test leaderboard. ${\spadesuit}$ denotes agent that utilizes graph information during navigation.
    }
    \vspace{-5pt}
    \label{table0}
\end{table*}

\paragraph{Masked Trajectory Modeling (MTM).}
In Masked Trajectory Modeling (MTM) (Top Left in Fig.~\ref{figure2}), we randomly mask out $r\%$ of the steps in the full navigation trajectory, and aim to recover those views based on language instructions and other views in the trajectory. Specifically, given language instruction $\{w_1,...,w_L\}$ and navigation trajectory $\{v_1,...,v_M\}$, we randomly replace some steps $v_i$ with a $\langle MASK \rangle$ token. We predict $\hat{p}^o_{ik_{MTM}}$ ($\hat{p}^o_{ik_{MTM}}\in \{\hat{p}^{o_m}_{ik_{MTM}}, \hat{p}^{o_s}_{ik_{MTM}}, \hat{p}^{o_w}_{ik_{MTM}}\}$) with a multi-layer fully-connected network: $\hat{p}^o_{ik_{MTM}} = f_{MTM}(h_i + E_p)$,
where $h_i$ is the model output of the masked step $v_i$. $E_p$ is zeros when we use mean patch probability, $E_p$ is the position encoding of the sampled patch when using sample patch probability, and $E_p$ is encoded sampled weights when using weighted patch probability. We optimize the KL-divergence loss between the two distributions:

\begin{equation}
    L_{MTM} = -\sum_{k\in S}{p^o_{ik}log\hat{p}^o_{ik_{MTM}}}
\end{equation}

\paragraph{Masked Panorama Modeling (MPM).}

At each time step, the agent observes a panorama of the current location. The panorama is discretized into 36 views. In Masked Panorama Modeling (Top Right in Fig.~\ref{figure2}), we randomly mask out $u$\% of the views in the panorama, and aim to predict the masked views based on language instructions, navigation history and surrounding views in the panorama. Specifically, given language instruction $\{w_1,...,w_L\}$, navigation history $\{v_1,...,v_M\}$, and current panorama view $\{v_{{M+1}_i}\}_{i=1}^{36}$, we randomly replace some views $v_{{M+1}_i}$ with a $<$MASK$>$ token. We predict $\hat{p}^o_{ik_{MPM}}$ with a multi-layer fully-connected network: $\hat{p}^o_{ik_{MPM}} = f_{MPM}(o_i + E_p)$,
where $o_i$ is the output of the masked discretized view. We optimize the KL-divergence loss between two distributions:

\begin{equation}
    L_{MPM} = -\sum_{k\in S}{p^o_{ik}log\hat{p}^{o}_{ik_{MPM}}}
\end{equation}

\paragraph{Action Prediction with Image Generation (APIG).}
Previous work picks the next action of the agent by directly selecting from a set of candidates. In this task, the agent mimics the action prediction process by generating the future scenes and picking the candidates that are closest to the generated images. In Action Prediction with Image Generation, we aim to generate semantics of the views at the next step, based on language instruction $\{w_1,...,w_L\}$ and navigation history $\{v_1,...,v_M\}$. We predict $\hat{p}^o_{ik_{APIG}}$ with a multi-layer fully connected network: $\hat{p}^o_{ik_{APIG}} = f_{APIG}(w_0 + E_p)$,
where $w_0$ is the output of the $<$CLS$>$ token in the instructions. The $<$CLS$>$ token contains state information of the agent after cross-modality attention layers between the instructions and the navigation history. We optimize the KL-divergence loss between two distributions:
\begin{equation}
    L_{APIG} = -\sum_{k\in S}{p^o_{ik}log\hat{p}^o_{ik_{APIG}}}
\end{equation}

\subsection{Fine-tuning on Navigation Task} \label{sec:ft}

We fine-tune the agent on the navigation task with a mixture of imitation learning (IL) and reinforcement learning (RL) as in~\cite{chen2021history}. To fully utilize agents' ability to generate semantics in future scenes, we further train the agent with APIG as an auxiliary task.
We initialize the model weights for this task with $f_{APIG}$ learned in pre-training. Since it's computationally inefficient to compute the ground truth generation probability for an image during VLN fine-tuning, we pre-extract the mean patch probability $p^{o_m}_{ik}$ of an image $v_i$ and the target visual token ($m_{ij} = argmax_k(p_{ijk})$) for every patch $j$ in an image $v_i$.
For $p_{ik}^{o_m}$, we optimize the KL-divergence loss between two distributions: $L_{AT} = -\sum_{k\in S}{p_{ik}^{o_m} log\hat{p}^{o_m}_{ik}}$. For $p_{ik}^{o_s}$, since we don't pre-extract the visual token probability $p_{ijk}$, we optimize the cross-entropy loss between predicted generation probability $\hat{p}^{o_s}_{ik}$ and the sampled visual token class $m_{ij}$ instead: $L_{AT} = -m_{ij}log\hat{p}^{o_s}_{im_{ij}}$. For $p_{ik}^{w}$, we fix the weights $w_j$ to be $1/N$ to generate mean patch probability instead, and optimize the KL-divergence loss as: $L_{AT} = -\sum_{k\in S}{p_{ik}^{o_w} log\hat{p}^{o_w}_{ik}}$. 

\begin{table*}
    \centering
    \resizebox{1.35\columnwidth}{!}{
    \begin{tabular}{|c|ccc|cccccc|}
    \hline 
     \textbf{No.}  & \multicolumn{3}{c|}{\textbf{Pre-Training Tasks}}  &
       \multicolumn{6}{c|}{\textbf{Validation Unseen}}\\  \hline 
       & \textbf{MTM} & \textbf{MPM} & \textbf{APIG}  & \textbf{NE$\downarrow$} & \textbf{SR$\uparrow$} & \textbf{SPL$\uparrow$} & \textbf{nDTW$\uparrow$} & \textbf{sDTW$\uparrow$}  & \textbf{CLS$\uparrow$} \\ \hline
1 & \xmark & \xmark & \xmark & - & 64.4 & 58.8 & - & - & - \\ \hline 
2 & \cmark & \xmark & \xmark & 3.52 & 66.7 & 61.3 & 68.7 & 57.3 & 67.3 \\
3 & \xmark & \cmark & \xmark & 3.68 & 65.5 & 59.4 & 68.2 & 56.5 & 66.8 \\ 
4 & \xmark & \xmark & \cmark & 3.58 & 66.3 & 60.9 & 68.6 & 57.2 & 67.0 \\ \hline
5 & \cmark & \cmark & \cmark &  \textbf{3.37} & \textbf{68.1} & \textbf{62.3}  &  \textbf{69.6} & \textbf{58.7} & \textbf{67.7} \\
    \hline
    \end{tabular}
    }
    \vspace{-5pt}
    \caption{Ablation results for our proposed three proxy tasks on Room-to-Room validation unseen set.
    }
    \vspace{-10pt}
    \label{table1}
\end{table*}

\begin{table*}
    \centering
    \resizebox{1.5\columnwidth}{!}{
    \begin{tabular}{|c|c|cc|cccccc|}
    \hline 
     \textbf{No.}  & \textbf{PT Tasks} & \multicolumn{2}{c|}{\textbf{Codebook Selection}}   &
       \multicolumn{6}{c|}{\textbf{Validation Unseen}}\\  \hline 
       &  & \textbf{Static} & \textbf{Dynamic}  & \textbf{NE$\downarrow$} & \textbf{SR$\uparrow$} & \textbf{SPL$\uparrow$} & \textbf{nDTW$\uparrow$} & \textbf{sDTW$\uparrow$} & \textbf{CLS$\uparrow$}  \\ \hline
1 & Baseline & & & - &64.4 & 58.8 & - & - & -  \\ \hline
2 & MTM & \xmark & \xmark & 3.66 & 65.1 & 59.8 & 68.5 & 56.7 & 66.5 \\ 
3 & MTM & \cmark & \xmark & 3.65 & 66.2 & 60.3 & 67.5 & 56.7 & 66.3 \\
4 & MTM & \xmark& \cmark  & 3.52 & 66.7 & 61.3 & 68.7 & 57.3 & 67.3 \\ \hline
5 & MPM & \xmark & \xmark & 3.89 & 64.8 & 59.0 & 68.0 & 55.4 & 65.3 \\
6 & MPM & \cmark & \xmark & 3.68 & 65.5 & 59.4 & 68.2 & 56.5 & 66.8  \\
7 & MPM & \xmark & \cmark &  3.64 & 65.1 & 59.8 & 68.0 & 56.3 & 66.2 \\ \hline
8 & APIG & \xmark & \xmark & 3.87 & 64.4 & 59.2 & 67.1 & 55.4 & 65.6   \\ 
9 & APIG & \cmark & \xmark &  3.76 & 65.9 & 60.3 & 67.4 & 56.1 & 65.6 \\
10 & APIG & \xmark & \cmark &  3.58 & 66.3 & 60.9 & 68.6 & 57.2 & 67.0 \\  
    \hline
    \end{tabular}
    }
    \vspace{-5pt}
    \caption{Comparison of training without codebook selection, training with static codebook selection method and dynamic codebook selection method on Room-to-Room validation unseen set.
    }
    \label{table2}
    \vspace{-3pt}
\end{table*}

\section{Experimental Setup}

\subsection{Dataset}
We evaluate our agent on the Room-to-Room (R2R) dataset \cite{anderson2018vision} and Cooperative Vision-and-Dialog Navigation (CVDN) dataset \cite{thomason2020vision}. Both datasets are split into training, seen validation, unseen validation and test set. The environments in unseen validation and test set are not appeared in the training set. We focus on agents' performance on the unseen validation set and the test set, since they measure agents' ability to generalize to unseen environments.

\subsection{Evaluation Metrics}
We evaluate our model on the following metrics: (1) Success Rate (SR), whether the agent stops within 3 meters to the target. (2) Success Rate Weighted by Path Length (SPL)~\cite{anderson2018evaluation}, which penalizes long paths that randomly explore the environment. (3) normalized Dynamic Time Warping (nDTW)~\cite{ilharco2019general}, measures how well the agent follows the ground truth path. (4) success rate weighted by normalized Dynamic Time Warping (sDTW), which considers nDTW score for success cases. (5) Coverage weighted by Length Score (CLS)~\cite{jain2019stay}, measures how well the agent follows the reference path. (6) Navigation Error (NE), the distance between the stop location and the target. (7) Trajectory Length (TL), the total navigation length of the agent. We consider success rate as the main evaluation metric.

\subsection{Implementation Details}
We adopt the model architecture from \cite{chen2021history}. For the image tokenizer, the input image size is 224, and the patch size is 16. We set 1.0 for $\gamma$ and 0.5 for $\lambda$ in dynamic codebook selection, and $|S|$ to be 1000 for both codebook selection methods. In pre-training, the ratio to select tasks is set to be 3 for MTM and 1 for others. The mask ratio $r$ is 0.5 for MTM and $u$ is 0.3 for MPM. In fine-tuning, $L_{AT}$ is added to the IL loss with ratio 1. The ratio to combine IL and RL is 0.15 when adding $L_{AT}$ and 0.2 otherwise. We use weighted patch probability with block-wise sampling for Room-to-Room dataset and base weighted patch probability for CVDN dataset. CLIP-ViT/16~\cite{radford2021learning} is used to extract the features for HAMT and ViT~\cite{dosovitskiy2020image} pretrained on ImageNet is used for DUET (CLIP-ViT/16 results for DUET are in Appendix). Other hyperparameters are the same as in \cite{chen2021history} for fair comparison.

\section{Results and Analysis}

\subsection{Test Set Results}
We show our method's performance on both the Room-to-Room (R2R) and the Cooperative Vision-and-Dialog Navigation (CVDN) dataset. 
As shown in Table~\ref{table0}, our agent outperforms previous SotA agent HAMT~\cite{chen2021history} which utilizes computational expensive end-to-end visual backbone tunning by 0.25 in GP on CVDN test leaderboard, and shows competitive performance on R2R test leaderboard. This demonstrates that our agent benefits from imagining the future view semantics during navigation. Furthermore, we demonstrate that our proposed methods generalize well to graph-based VLN agent DUET~\cite{chen2022think}, where the agent encodes the current location with neighboring nodes representation and makes action conditioned on both coarse-scale topological map encoding and fine-scale current step encoding. Specifically, we pre-train the agent with Masked Pano Modeling (MPM) and Action Prediction with Image Generation (APIG). Our method outperforms DUET~\cite{chen2022think} by 3\% in success rate and 1\% in SPL, achieving the new state-of-the-art on Room-to-Room test leaderboard.

\begin{table*}
    \centering
    \resizebox{1.8\columnwidth}{!}{
    \begin{tabular}{|c|c|cccc|cccccc|}
    \hline 
     \textbf{No.}  & \textbf{PT Tasks} & \multicolumn{4}{c|}{\textbf{Semantic Calculation}}   &
       \multicolumn{6}{c|}{\textbf{Validation Unseen}}\\  \hline 
       &  & \textbf{Mean} & \textbf{Sample} & \textbf{Weighted} & \textbf{Weighted-Block}  & \textbf{NE$\downarrow$} & \textbf{SR$\uparrow$} & \textbf{SPL$\uparrow$} & \textbf{nDTW$\uparrow$} & \textbf{sDTW$\uparrow$} & \textbf{CLS$\uparrow$}  \\ \hline
1 & Baseline & & & & &- &64.4 & 58.8 & - & - & -  \\ \hline
2 & ALL  & \cmark & \xmark & \xmark & \xmark & 3.66 & 65.5 & 60.3 & 68.8 & 56.4 & 67.4 \\
3 & ALL & \xmark & \cmark & \xmark & \xmark & 3.67 & 65.9 & 60.2 & 68.0 & 56.6 & 66.5 \\ 
4 & ALL & \xmark & \xmark & \cmark & \xmark& 3.43 & 66.9 & 61.8 & \textbf{69.7} & 58.2 & \textbf{68.4} \\
5 & ALL & \xmark & \xmark & \xmark & \cmark &  \textbf{3.37} & \textbf{68.1} & \textbf{62.3}  &  69.6 & \textbf{58.7} & 67.7 \\ 
    \hline
    \end{tabular}
    }
    \vspace{-5pt}
    \caption{Comparison of different ways to calculate image semantics on Room-to-Room validation unseen set. ``Weighted" indicates using weighted patch probability without block-wise sampled weights, and ``Weighted-Block" samples block-wise weights when calculating weighted patch probability.
    }
    \vspace{-10pt}
    \label{table3}
\end{table*}

\begin{table*}
    \centering
    \resizebox{1.6\columnwidth}{!}{
    \begin{tabular}{|c|c|cc|c|cccccc|}
    \hline 
     \textbf{No.}  & \textbf{PT Tasks} & \multicolumn{2}{c|}{\textbf{Codebook Selection}}   & \multicolumn{1}{c|}{\textbf{FT Tasks}}   &
       \multicolumn{6}{c|}{\textbf{Validation Unseen}}\\  \hline 
       &  & \textbf{Static} & \textbf{Dynamic}  &
       \textbf{APIG} & \textbf{NE$\downarrow$} & \textbf{SR$\uparrow$} & \textbf{SPL$\uparrow$} & \textbf{nDTW$\uparrow$} & \textbf{sDTW$\uparrow$} & \textbf{CLS$\uparrow$}  \\ \hline
1 & Baseline & & & & - &64.4 & 58.8 & - & - & -  \\ \hline
2 & APIG & \cmark & \xmark & \xmark & 3.64 & 64.9 & 60.4 & 67.9 & 56.2 & 66.7 \\ 
3 & APIG & \cmark & \xmark & \cmark & 3.76 & 65.9 & 60.3 & 67.4 & 56.1 & 65.6  \\
4 & APIG & \xmark & \cmark &  \xmark & 3.53 & 66.1 & 60.7 & 68.1 & 56.9 & 66.5 \\ 
5 & APIG & \xmark & \cmark & \cmark & 3.58 & 66.3 & 60.9 & 68.6 & 57.2 & 67.0 \\
    \hline
    \end{tabular}
    }
    \vspace{-5pt}
    \caption{Ablation results for adding Action Prediction with Image Generation (APIG) as auxiliary task during fine-tuning on Room-to-Room validation unseen set.
    }
    \vspace{-10pt}
    \label{table4}
\end{table*}

\subsection{Effectiveness of Pre-training Tasks}
In this section, we demonstrate the effectiveness of our proposed three tasks for VLN in-domain pre-training. We show the ablation performance of using one of the tasks in pre-training in Table~\ref{table1}. 
Specifically, the baseline model is HAMT pre-trained on 6 proxy tasks: MLM, MRM, ITM, SAP/SAR, and SPREL without end-to-end vision-backbone tunning (details in \cite{chen2021history}). The performance shown in this table is the best model picked among the different combinations of codebook selection methods and patch probability calculation methods. As shown in Table~\ref{table1}, adding any of our proposed tasks improves the performance by a large margin in both success rate (SR) and success rate weighted by path length (SPL). Specifically, adding Masked Trajectory Modeling (MTM) brings the largest improvement, outperforms the baseline by 2.3\% in SR and 2.5\% in SPL. Furthermore, pre-training on all the tasks achieves the best performance, improving the baseline by 3.7\% in SR and 3.5\% in SPL, demonstrating our approach's effectiveness.

\subsection{Effectiveness of Codebook Selection}

We demonstrate that it's important to select a subset of visual tokens either statically or dynamically for the agent to effectively learn the large visual token vocabulary extracted with pre-trained dVAE. We show the ablation performance in Table~\ref{table2}. First, we observe that the agent cannot learn useful image semantics when training on MTM or APIG without codebook selection (Model 1 vs. 2,8), but using dynamic codebook selection could significantly improve the performance, outperforming the model without codebook selection by 1.6\% in SR for MTM (Model 2 vs. 4) and 1.9\% in SR for APIG (Model 8 vs. 10). For MPM, using static codebook selection slightly outperforms the model that directly learns from all visual tokens in the vocabulary. We attribute this to that reasoning the masked view in a panorama is easier than generating the semantics in a masked step or next view. The agent could easily infer semantics from the overlaps between continuous views, and thus codebook selection benefits less for MPM.

\subsection{Comparison of Image Semantic Representation Methods}

We compare the three ways of representing the overall image semantics in Table~\ref{table3}. First, we observe that the agent benefits from pre-training with our proposed three tasks, regardless of how we calculate the image semantics (Model 1 vs. 2,3,4,5). Then, we observe that using sample patch probability, weighted patch probability without block-wise sampling, and weighted patch probability with block-wise sampling outperforms the mean patch probability (Model 2 vs. 3,4,5) by more than 0.5\%, 1.4\%, and 2.6\% in SR respectively. This demonstrates that our proposed sample patch probability and weighted patch probability (with and without block-wise sampling) help the agent learn from semantics in each patch better, while saving the memory by $O(N)$ compared with traditional Masked Image Modeling method~\cite{bao2021beit} which generates the probability for all masked patches in an image. Lastly, we demonstrate that adding block-wise sampling for the weights during training (Model 4 vs. 5) further improves the performance by 1.2\% in success rate, demonstrating the importance of learning semantics in a block instead of sparsely at each optimization step.

\subsection{Effectiveness of Auxiliary Task during Fine-tuning}
We demonstrate the effectiveness of utilizing future semantics generation as auxiliary task during fine-tuning in this section. We use mean patch probability to represent the image semantics in these experiments. As shown in Table~\ref{table4}, adding APIG as auxiliary task during fine-tuning improves the performance by 1.0\% in SR when using static codebook selection, and 0.2\% in SR when using dynamic codebook selection. This shows that the agent can benefit from generating semantics in future views during navigation. 

\begin{figure}[t]
\begin{center}
\includegraphics[width=0.9\linewidth]{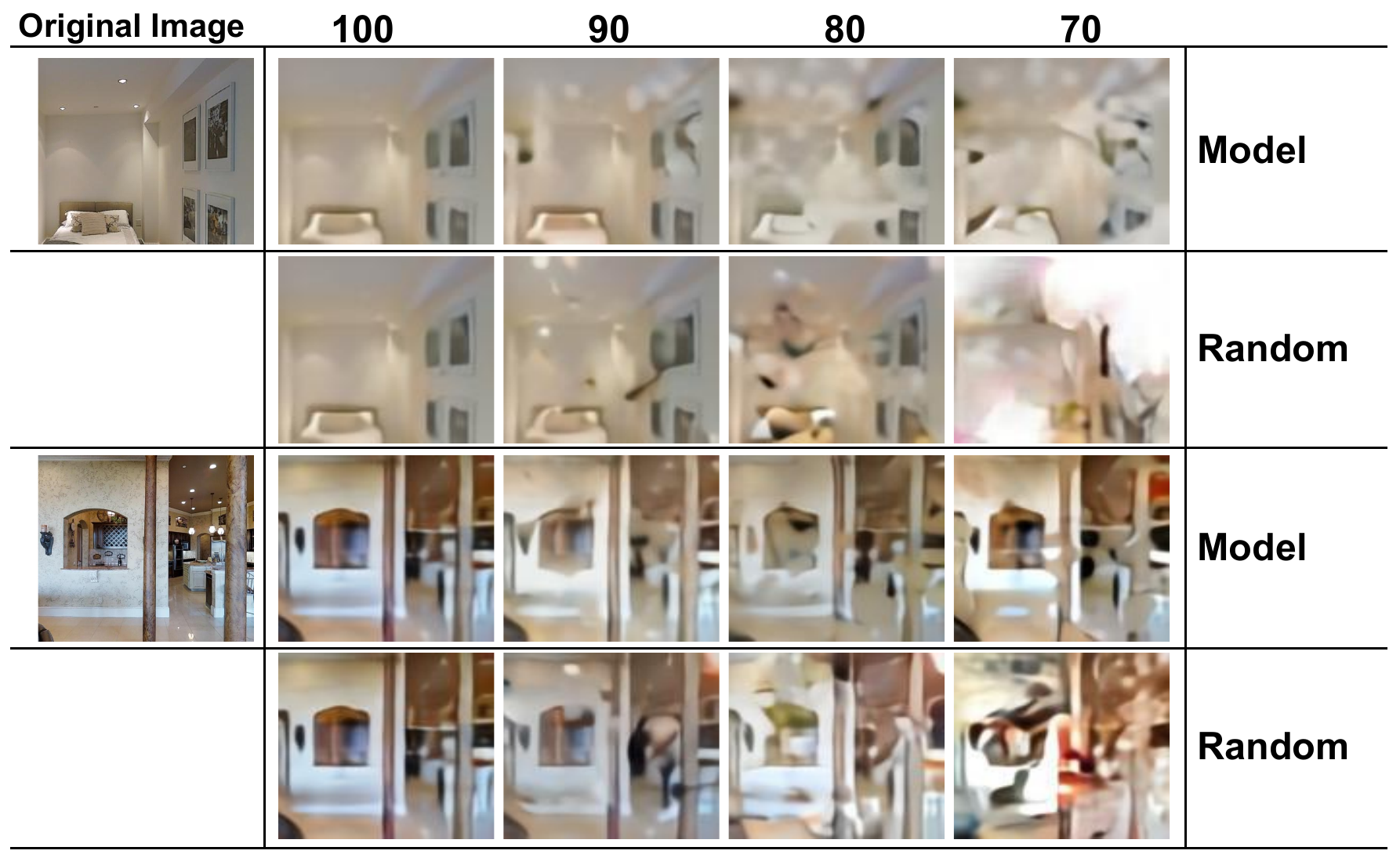}
\end{center}
\vspace{-10pt}
  \caption{Qualitative analysis of images generated with our agent. Given 90\%/ 80\%/ 70\% ground truth tokens, the rest tokens predicted by our model contain closer semantic information compared with randomly filled tokens.
 }
 \vspace{-10pt}
 \label{figure3}
\end{figure}

\subsection{Ability to Generate Future Views}
We demonstrate that our model could reasonably generate future semantics and reconstruct future images. We generate the token for each patch in the image with our APIG head learned with weighted patch probability. We set the weight vector to be an indicator vector, where the weight for the target patch is 1 and the others are 0. We use $x$\% ($x\in{70, 80, 90}$) of the patches with ground truth tokens encoded with pre-trained dVAE, the rest patches with the tokens generated by our APIG head, and then use the dVAE decoder to decode the image. As shown in Figure~\ref{figure3}, our generated image could almost reconstruct the beds and pictures in the original image with small vague areas when given 70\% of the ground truth tokens. In comparison, filling the 30\% patches with random tokens will generate distorted images with large white parts, and the beds and pictures information cannot be identified. We also notice that our model still fails to generate the full image when all the tokens in the image are predicted by our APIG head, and needs at least 70\% of the ground truth tokens to generate images of reasonable quality. More examples in appendix.

\begin{table}
    \centering
    \resizebox{0.75\columnwidth}{!}{
    \begin{tabular}{|c|cccc|}
    \hline 
     \textbf{Path Length}   &  \textbf{3} & \textbf{4} & \textbf{5}   & \textbf{6}  \\ \hline
    HAMT~\cite{chen2021history} & 66.7 & 65.8 & 67.5 & 62.3 \\
    Ours & 66.7 & \textbf{67.7} & \textbf{70.3} & \textbf{66.5} \\
    \hline
    \end{tabular}
    }
    \vspace{-5pt}
    \caption{Success rate of our method and the baseline method for navigation trajectories with different lengths on Room-to-Room validation unseen set. Our method achieves larger improvement for longer paths.
    }
    \vspace{-13pt}
    \label{table6}
\end{table}

\subsection{Better Performance for Longer Trajectories}
We show that our model is better at navigating longer paths on validation unseen set. Room-to-Room validation unseen set contains path lengths ranging from 3 to 6. As shown in Table~\ref{table6}, our model improves the baseline by 4.2\% in success rate for paths with length 6, 2.8\% for paths with length 5, and 1.9\% for paths with length 4, demonstrating that learning to predict future view semantics can help the agent to learn to navigate longer paths better.

\section{Conclusion}
In this paper, we propose \methodname{}, which explores whether navigation agents can benefit from imagining semantics in future views. We first propose novel codebook selection and image semantic calculation methods to efficiently learn the image semantics. We then design Masked Trajectory Modeling and Masked Panorama Modeling to help the agent to understand semantics in each visual token and learn to recognize the semantics contained in each view. We further use Action Prediction with Image Generation to enhance agents' ability to predict future views in both pre-training and fine-tuning. Our \methodname{} achieves the new SotA on both CVDN dataset and R2R dataset. 

\section*{Acknowledgements}
We thank the reviewers for their helpful comments. This work was supported by ARO Award W911NF2110220 and DARPA MCS N66001-19-2-403. The views contained in this article are of the authors and not of the funding agency.

{\small
\bibliographystyle{ieee_fullname}
\bibliography{egbib}
}

\appendix
\section{Appendix Overview}
In this appendix, we provide the following:
\begin{itemize}
    \item Detailed description of the architecture of the baseline method we use in Sec.~\ref{hamt}.
    \item Detailed description of the pre-training tasks used in the baseline method in Sec.~\ref{hamt_pt}, and more implementation details in Sec.~\ref{implementation}.
    \item Performance of our method on R4R and RxR validation unseen set in Sec.~\ref{r4rrxr}, and performance of our method on graph based navigation agents on R2R validation set in Sec.~\ref{duetclip}.
    \item Proof that demonstrates weighted patch probability could learn a better optimal value than mean patch probability in Sec.~\ref{proof}. 
    \item More examples of the future views generated by our APIG head in Sec.~\ref{generation}, and quantitative analysis of the semantics underlying our generated tokens in Sec.~\ref{semantic_analysis}.
\end{itemize}

\section{HAMT Model Architecture} \label{hamt}
HAMT~\cite{chen2021history} utilizes a transformer-based architecture to encode the instructions, navigation history, and current step observation. Specifically, the instructions are encoded with a BERT architecture. 

As the navigation history is a sequence of panorama observations, HAMT encodes the navigation history with a hierarchical architecture. It first uses a panorama encoder to encode panorama into view representation, and uses multiple transformer layers as the temporal encoder to encode the observations on the trajectory. Formally, given the encoded history observation $v_i$, which is the output of the panorama encoder, the output of the temporal encoder is $h_i = LN(W_tv_i) + LN(W_aa_i) + E_i^S + E_2^T$, where $a_i$ is the action embedding at step $i$, $E_i^S$ is the step encoding, and $E_2^T$ is the token type encoding which indicates the input is history views. 

The current step observation is represented as 36 discretized views. Each view is passed through the transformer encoder to learn the view representation: $o_i = LN(W_ov_i^o) + LN(W_a^o a_i^o) + E_i^O + E_1^T$, where $v_i^o$ is the encoding of view $i$, $a_i^o$ is the action embedding for view $i$, $E_i^O$ is the embedding indicating whether the current view is navigable, and $E_1^T$ is the token type encoding which indicates the input is current step observation.

The agent predicts the next step by comparing the similarity between the observation encoding $o_i$ and the $<$CLS$>$ token which contains instruction-trajectory information. 

More implementation details can be found in \cite{chen2021history}.

\section{Pre-training Tasks in HAMT} \label{hamt_pt}
In this section, we describe the six pre-training tasks we adopted from ~\cite{chen2021history}.
Specifically, the six tasks are Masked Language Modeling (MLM), Masked Region Modeling (MRM), Instruction Trajectory Matching (ITM), Single-step Action Prediction/Regression (SAP/SAR), and Spatial Relationship Prediction (SPREL). 

In Masked Language Modeling, we randomly masked out 15\% of the words in the instructions, and predict the masked words given surrounding words and the full trajectory, which improves agents' language understanding. We optimize the negative log-likelihood of the original words: $L_{MLM} = -logp(w_i | I_{lang\setminus{i}}, I_{visual})$, where $I_{lang\setminus{i}}$ is the language input without masked words $w_i$ and $I_{visual}$ is the visual input.

In Masked Region Modeling, the agent learns to predict the objects in the masked views in the trajectory. The target is the object detection probability $p_i$ predicted by an image classification model pre-trained on ImageNet. We optimize a KL-divergence between the target probability and predicted probability: $L_{MRM} = -p_ilog\hat{p}_i$, where $\hat{p}_i$ is the predicted probability.

In Instruction Trajectory Matching, the agent learns the alignment between the language instructions and the environment by picking the correct instruction-trajectory pairs from one positive pair and four negative pairs. The four negative pairs are created by randomly sampling two trajectories from the same batch, and shuffling the order of views in the correct trajectory. The agent optimizes a noisy contrastive loss: $L_{ITM} = -log \frac{exp(f_{ITM}(h_{lang} * h_{visual}))}{exp(f_{ITM}(h_{lang} * h_{visual})) + \sum^4{exp(f_{ITM}(h_{lang} * h_{visual}^{neg}))}}$, where $h_{lang}$ and $h_{visual}$ are the outputs of the $<$CLS$>$ token of the instructions and the trajectories separately. 

In Single-step Action Prediction and Single-step Action Regression, the agent needs to select the next step from a set of candidates. Specifically, the agent optimizes a negative log probability of the target view action in Single-step Action Prediction, and predicts the heading and elevation of the target view action by optimizing the L2 loss. 

In Spatial Relationship Prediction, the agent learns to predict the relative spatial position of two views in a panorama. Specifically, it optimizes a L2 loss between the predicted heading and elevation difference and ground truth heading and elevation difference between two views. 

More implementation details can be found in \cite{chen2021history}.

\begin{figure}[t]
\begin{center}
\includegraphics[width=1.0\linewidth]{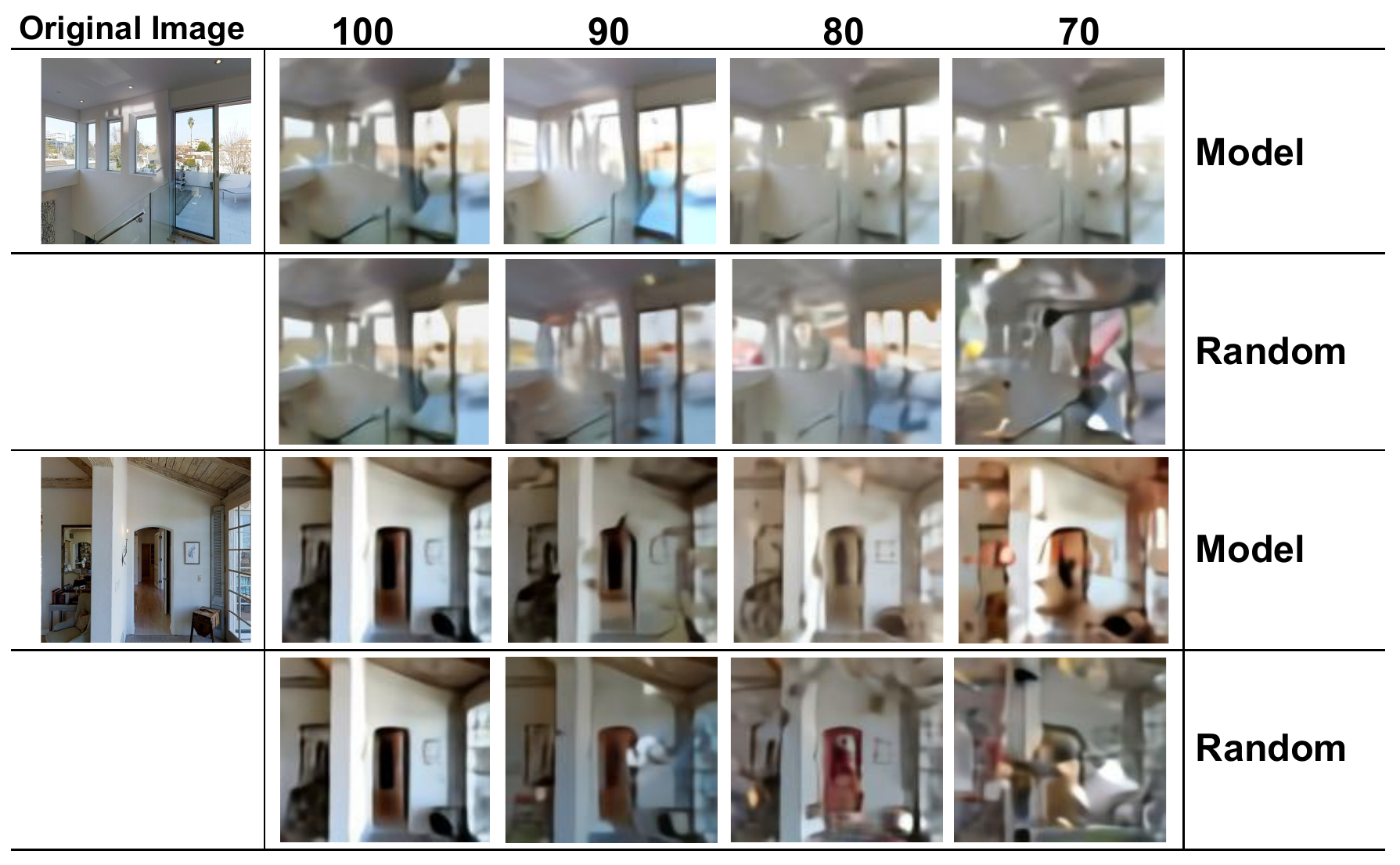}
\end{center}
  \caption{Qualitative analysis of images generated with our agent. Given 90\%/ 80\%/ 70\% ground truth tokens, the rest tokens predicted by our model contain closer semantic information compared with randomly filled tokens.
 }
 \label{figure1_appendix}
\end{figure}

\section{Implementation Details} \label{implementation}
We adopt the model architecture from \cite{chen2021history}. For the image tokenizer, the input image size is 224, and the patch size is 16. We set 1.0 for $\gamma$ and 0.5 for $\lambda$ in dynamic codebook selection, and $|S|$ to be 1000 for both codebook selection methods. In pre-training, the ratio to select tasks is set to be 3 for MTM and 1 for others. The mask ratio $r$ is 0.5 for MTM and $u$ is 0.3 for MPM. In fine-tuning, $L_{AT}$ is added to the IL loss with ratio 1. The ratio to combine IL and RL is 0.15 when adding $L_{AT}$ and 0.2 otherwise. We use weighted patch probability with block-wise sampling for Room-to-Room dataset and base weighted patch probability for CVDN dataset. Other hyperparameters are the same as in \cite{chen2021history} for fair comparison. For training time, our model takes 30 hours to converge on 2 NIVIDIA A6000 GPU, while HAMT full model takes 20 hours training on 20 NVIDIA V100 GPUs in addition to initial pre-training for 1 day on 4 NVIDIA Tesla. For model parameters updated during pre-training, our model updates 191M parameters while HAMT full model updates 260M, saving 27\% parameters. 

\begin{table}
    \centering
     \resizebox{1.0\columnwidth}{!}{
    \begin{tabular}{|c|cccc|ccc|}
    \hline 
     \textbf{Model}  & \multicolumn{4}{c|}{\textbf{R4R}}  &
       \multicolumn{3}{c|}{\textbf{RxR}}\\  \hline 
     &  \multicolumn{4}{c|}{\textbf{Val Unseen}} &  \multicolumn{3}{c|}{\textbf{Val Unseen}} \\ \hline
       & \textbf{SR$\uparrow$}  & \textbf{nDTW$\uparrow$} & \textbf{sDTW$\uparrow$} & \textbf{CLS$\uparrow$} & \textbf{SR$\uparrow$} & \textbf{nDTW$\uparrow$} & \textbf{sDTW$\uparrow$}  \\ \hline
HAMT~\cite{chen2021history}  & 44.6 & 50.3 & 31.8 & 57.7 & 56.5 & 63.1 & 48.3  \\
Ours &  \textbf{45.8} & \textbf{52.9} & \textbf{33.6} & \textbf{59.1} & \textbf{60.0} & \textbf{65.3} & \textbf{51.4} \\ \hline 
    \end{tabular}
    }
    \caption{Comparison with state-of-the-art agents on R4R and RxR validation unseen set. 
    }
    \label{table7}
\end{table}

\section{Performance on R4R and RxR Dataset} \label{r4rrxr}
In this section, we show our agents' performance on R4R and RxR dataset. As shown in Table~\ref{table7}, our model surpasses the HAMT full model by 2.6\% in nDTW and 1.8\% in sDTW on R4R validation unseen set. Besides, we show that our model achieves significantly better performance on RxR unseen set, improving the HAMT method by 2.1\% in nDTW and 3.1\% in sDTW.

\begin{table}
    \centering
     \resizebox{1.0\columnwidth}{!}{
    \begin{tabular}{|c|cccc|cccc|}
    \hline 
     \textbf{Model} 
     &  \multicolumn{4}{c|}{\textbf{Val Seen}} &  \multicolumn{4}{c|}{\textbf{Val Unseen}} \\ \hline
       & \textbf{SR$\uparrow$}  & \textbf{SPL$\uparrow$} & \textbf{nDTW$\uparrow$} & \textbf{sDTW$\uparrow$} & \textbf{SR$\uparrow$} & \textbf{SPL$\uparrow$} & \textbf{nDTW$\uparrow$} & \textbf{sDTW$\uparrow$} \\ \hline
DUET-CLIP~\cite{chen2022think}  & 75.2 & 69.1 & 76.0 & 66.5 & \textbf{72.8} & 63.4 & 69.3 & 60.6  \\
Ours &  \textbf{78.6} & \textbf{74.8} & \textbf{81.8} & \textbf{73.2} & 72.2 & \textbf{63.7} & \textbf{70.2} & \textbf{61.4} \\ \hline 
    \end{tabular}
    }
    \caption{Comparison with state-of-the-art agents on R2R validation set with CLIP-ViT/16 features. 
    }
    \label{table8}
\end{table}

\section{Performance with DUET on R2R Dataset with CLIP Features} \label{duetclip}
We also show the performance of our proposed methods when generalized to graph-based VLN agent DUET~\cite{chen2022think} with CLIP-ViT/16 features (instead of ViT features as in the main paper). Specifically, we pre-train the agent with Masked Pano Modeling (MPM) and Action Prediction with Image Generation (APIG). As shown in Table~\ref{table8}, when using CLIP-ViT/16 features, our method outperforms DUET~\cite{chen2022think} by 3.4\% in success rate and 5.7\% in SPL in validation seen environment, and 0.3\% in success rate and 0.9\% in nDTW in validation unseen environment, demonstrating the effectiveness of our approach.

\section{Weighted Patch Probability Proof} \label{proof}
In weighted patch probability calculation, we represent image semantics generation probability $p^{o_w}_{ik}$ as:
\begin{equation}
    p^{o_w}_{ik} = \sum_{j=1}^{N}{w_jp_{ijk}} 
\end{equation}
where $w_j$ is a randomly sampled weight for patch $j$, and $\sum_{j=1}^N{w_j}=1$. 

During training, we randomly sample $\{w_j\}_{j=1}^N$ for each example, and minimize the differences between $\sum_{j=1}^{N}{w_jp_{ijk}}$ and predicted $\hat{p}^{o_w}_{ik}$. We hypothesize that the agent learns to predict $\hat{p}^{o_w}_{ik}$ by predicting $\sum_{j=1}^{N}{w_j\hat{p}_{ijk}}$ as we conditioned the prediction based on sampled weights $\{w_j\}_{j=1}^N$.

We want to reach the optimal where:
\begin{equation}
p^{o_w}_{ik} = \hat{p}^{o_w}_{ik}
\end{equation}

\begin{equation}
    \sum_{j=1}^{N}{w_jp_{ijk}} = \sum_{j=1}^{N}{w_j\hat{p}_{ijk}} 
\end{equation} 

Since we randomly sample $\{w_j\}_{j=1}^N$ during training, this guarantees that for every patch $j$ in the image $v_i$, $p_{ijk} = \hat{p}_{ijk} $, otherwise $\exists \{w_j\}_{j=1}^N$ that makes  $\sum_{j=1}^{N}{w_jp_{ijk}} \neq \sum_{j=1}^{N}{w_j\hat{p}_{ijk}}$.

\section{Future View Generation Examples} \label{generation}
We demonstrate that our model could reasonably generate future semantics and reconstruct future images with more examples. As shown in Figure~\ref{figure1_appendix}, our generated image could almost reconstruct the doors and the overall layout of the room when given 70\% of the ground truth tokens. In comparison, filling the 30\% patches with random tokens will produce distorted images which are hard to infer how does the original images look like.

\section{Analysis of Generated Semantics} \label{semantic_analysis}
In this section, we compare the generated semantics with the ground truth semantics quantitatively to demonstrate that the semantic information underlying them is similar. Specifically, we represent the semantics of each visual token as the output of the first embedding layer in the dVAE decoder (which maps each token to a 128 dimension representation space). We calculate the distance between generated semantics and ground truth semantics, and compare it with the distance between the ground truth semantics and all other tokens in the vocabulary (i.e., the distance between the ground truth token and other 8191 tokens for each patch). We normalize each semantic representation and use l2-norm as the distance. Our method has a distance of 0.95, while the baseline is 1.31. This shows that the distance between our generated semantics and ground truth semantics is closer.

\end{document}